\documentclass[superscriptaddress,twocolumn, amsmath,amssymb, aps,prx,citeautoscript]{revtex4-2}
\setcitestyle{super}

\usepackage{graphicx}
\usepackage{booktabs}
\usepackage{siunitx}
\usepackage{amsfonts}
\usepackage{amsmath}
\usepackage{bm}
\usepackage[capitalize]{cleveref}

\usepackage{caption}
\usepackage{subcaption}

\begin{document}

\title{CoordGate: Efficiently Computing Spatially-Varying Convolutions in Convolutional Neural Networks}

\author{Sunny Howard}
\email{sunny.howard@physics.ox.ac.uk}
\affiliation{Department of Physics, Clarendon Laboratory, University of Oxford, Parks Road, Oxford OX1 3PU, United Kingdom}%
\affiliation{Faculty of Physics, Ludwig--Maximilians--Universit{\"a}t M{\"u}nchen, Am Coulombwall 1, 85748 Garching, Germany}%

\author{Peter Norreys}
\affiliation{Department of Physics, Clarendon Laboratory, University of Oxford, Parks Road, Oxford OX1 3PU, United Kingdom}%
\affiliation{John Adams Institute for Accelerator Science, Denys Wilkinson Building, Oxford OX1 3RH, United Kingdom}%
\author{Andreas Döpp}
\email{a.doepp@physik.uni-muenchen.de}
\affiliation{Department of Physics, Clarendon Laboratory, University of Oxford, Parks Road, Oxford OX1 3PU, United Kingdom}%
\affiliation{Faculty of Physics, Ludwig--Maximilians--Universit{\"a}t M{\"u}nchen, Am Coulombwall 1, 85748 Garching, Germany}%
\email{a.doepp@lmu.de}

\begin{abstract}
Optical imaging systems are inherently limited in their resolution due to the point spread function (PSF), which applies a static, yet spatially-varying, convolution to the image. This degradation can be addressed via Convolutional Neural Networks (CNNs), particularly through deblurring techniques. However, current solutions face certain limitations in efficiently computing spatially-varying convolutions. In this paper we propose CoordGate, a novel lightweight module that uses a multiplicative gate and a coordinate encoding network to enable efficient computation of spatially-varying convolutions in CNNs. CoordGate allows for selective amplification or attenuation of filters based on their spatial position, effectively acting like a locally connected neural network. The effectiveness of the CoordGate solution is demonstrated within the context of U-Nets and applied to the challenging problem of image deblurring. The experimental results show that CoordGate outperforms conventional approaches, offering a more robust and spatially aware solution for CNNs in various computer vision applications. 
\end{abstract}

\maketitle

\section{Introduction}
\label{sec:intro}

Convolution is a fundamental operation that lies at the core of methods from numerous disciplines, from physical processes like heat transfer, to convolutional neural networks in machine learning. In an optics context, this operation involves sliding a point spread function (PSF) - a system's response to a point source - over an input signal in order to obtain the convolved signal. While classical convolution requires a PSF that is spatially-invariant (i.e. no change as it is passed over the input), this is a property that is rarely found in reality, due to, for example, optical aberrations. When one considers a convolution with a spatially-variant PSF, the expressability of the operation becomes even greater. However, with greater functionality, comes greater computational complexity. Discretised spatially-varying convolution is described by the general formula, 

\begin{equation}
    n(\mathbf{i}) = \sum _{\mathbf{j}\in\Omega(\mathbf{i})} h(\mathbf{i},\mathbf{j})m(\mathbf{j}),\label{SVconvolution}
\end{equation}
where $m$ and $n$ are the input and convolved signals, $h$ is the point spread function, $\mathbf{i}$ is a position, and $\Omega(\mathbf{i})$ describes a relevant region around $\mathbf{i}$ (which may include the full domain of $n$). Classic, spatially-invariant, convolution enforces that $h(\mathbf{i},\mathbf{j})=h(\mathbf{i}-\mathbf{j})$. 

When imaging an object, a PSF is known to be dependant on the in-plane spatial coordinates, $(x,y)$, as well as the  angles of incidence,  $\mathbf{\theta}$ (or equivalently depth, $z$). Using a typical single two-dimensional (2D) sensor does not allow access to the latter variable, which leads to deconvolution becoming underdetermined. Here, a \emph{static} spatially-varying convolution is defined as one where the PSF is consistent for every data sample and depends only on the in-plane coordinates, so that the other degree of freedom is removed. Such a condition is satisfied in many scenarios in optics; for example, when imaging at a fixed depth ($z(x,y)=C(x,y)$), such as in microscopy or in relay imaging, or at very far distances ($\theta\to0)$, such as in astronomy. The aforementioned definition also typically removes examples that include motion blur, which would not be static between data samples. Finally, one notes that the (pseudo-)inversion of a static convolution gives rise to a static spatially-varying deconvolution, which in a noise-free case is totally determined. This paper focuses on the problem of performing static spatially-varying convolution and deconvolution. 


Recent efforts in these problems have turned to deep learning methods, such as convolutional neural networks (CNNs). CNN's are composed of convolutional layers, which have the typically desirable characteristic of weight-sharing - significantly reducing the number of trainable parameters in the network and allowing for efficient position-independent extraction of local features from images. By sequentially stacking convolutional layers, CNNs extract highly abstract features, whilst also widening their receptive field.  However, the property of weight-sharing also imposes a constraint on the model's ability to learn spatially-aware representations. While CNNs can actually detect spatially varying features \cite{boundary_effects}, the method of how they do this is inefficient, as will be described in subsequent sections of this paper. Several methods have been suggested to improve CNN's spatial capabilities, one of which is the CoordConv layer \cite{coordconv}, which appends coordinates to the input features, enabling the network to learn spatial-aware representations.


In this paper, we revisit the problem of CNNs' inefficiency in spatial awareness and propose a novel solution, CoordGate. While CoordConv appends the coordinates to the data before convolution, in CoordGate they are passed through an encoder network, before being applied to the convolved data via a multiplication gate, similar to that found in a channel-attention mechanism. This technique enables selective amplification or attenuation of filters based on their spatial position, and it provides a large efficiency increase over existing CNNs. The paper is structured as follows: In Section II, a discussion is provided on the state of the art regarding spatially varying convolutions in CNN architectures and relevant related work is highlighted. This section serves as a basis for the subsequent introduction of the proposed solution in Section III and experimental results are provided in Section IV. Section V summarizes the new findings and concludes the paper.

\section{State of the Art and Related Work}\label{sec:related}

\begin{figure*}[!htb]
    \centering
    \includegraphics[width=\linewidth]{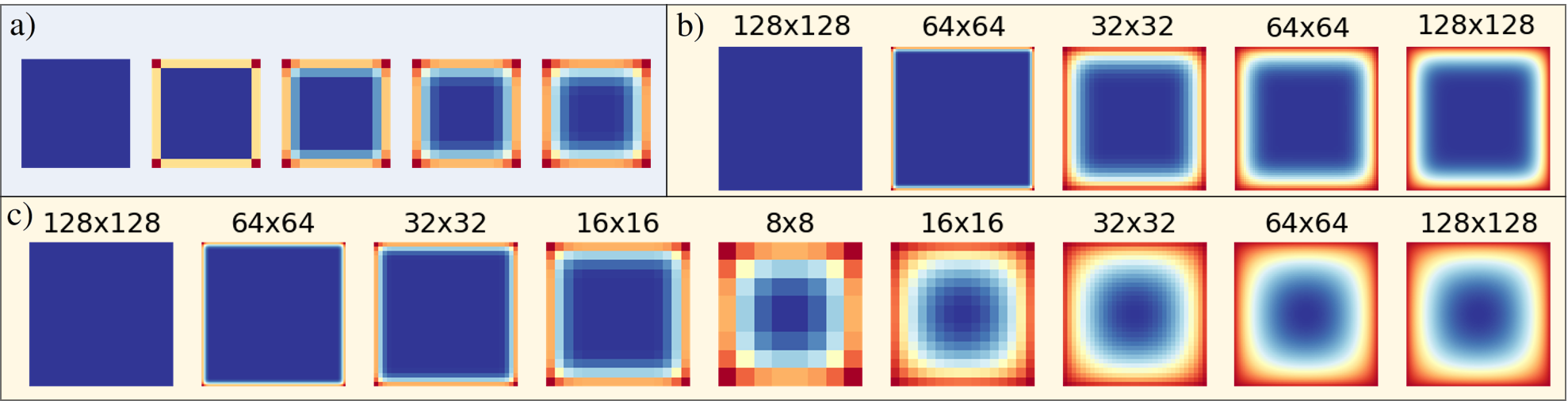}
    \caption{The position encoding effects from \emph{`same'} padding. (a): Convolving a uniform input with a $3\times 3$ uniform kernel 5 times. (b\&c): The same effect for 2 U-Net architectures, containing 2 and 4 steps of down-and-up sampling respectively. Before each dimension-changing operation, three $3\times3$ convolutions were applied, except for the middle layer in (b), where 12 additional convolutions were applied so each model had the same total number of convolutions.}
    \label{boundary}
\end{figure*}

\noindent \textbf{Boundary Effects}. Standard CNNs inadvertently leverage spatial variance due to padding effects \cite{boundary_effects,boundary_effects2}. It is common to pad the input to a convolutional layer with zeros before passing the kernel over it (\emph{`same'} padding), as this maintains the spatial size of the image. Post convolution, a systematic defect appears around the feature map's boundary, which seeps inwards with each subsequent convolutional layer. This effect is observable in \cref{boundary}(a), where a ($12\times12$) uniform input is convolved with a ($3\times3$) uniform kernel using `same' padding. The resultant feature map displays the position-encoding effect propagating inward. A different kernel would disrupt symmetry in these maps.
A network cannot detect spatial variance in a region until the defect reaches it; this is visible in the central region of \cref{boundary}((a) that retains a uniform value. In a standard computer vision task with an input size of $\sim 500\times 500$, a significant number of $3\times 3$ convolutions are needed to reach the center. Using encoder-decoder architectures like U-Net \cite{unet} can alleviate this issue. Convoluting at the downsampled layer in U-Net helps encode positional information in fewer steps, but at a downsampled resolution, as shown in \cref{boundary}(b-c).  Thus, for U-Net to exploit positional information, it is necessary for the network to be deep and it may require a large number of channels to describe more complicated positional relationships. This unintended positional encoding of padding contributes to deep CNN's performance in spatial tasks, but it encourages exploration of more deliberate encoding methods for potentially higher accuracy and efficiency.
\linebreak

\noindent\textbf{Adaptive Convolution.} It is possible to adapt a classic convolution in order to give it some spatial variance. One of the simplest methods in this class is the CoordConv layer, which concatenates normalized spatial coordinates to the input features before they are passed through a convolution, aiming to allow the network to learn spatially-aware representations more explicitly \cite{coordconv}, and it has been adopted for various applications \cite{wang2020solo,imageselfattention,sitzmann2019deepvoxels,el2021coordconv,svcnn,uselis2020localized}.  In the case of static spatially-varying convolution, one desires that the coordinate information affects the feature map in the same way for every data sample. Throughout the training set, the values of data will vary, while the value of the coordinates will be static. CoordConv layers include the coordinate within a weighted sum with the data values which makes it impossible to find weights for a convolutional kernel that will allow the coordinates to effect each sample in the same way. This is a fundamental limitation of CoordConv.

Another method trying to address this issue is the pixel-adaptive convolution (PAC) \cite{pixeladaptiveconvolution}. In this method, the actual convolution function is multiplied by a pairwise function of pixel features (such as the coordinates). However, in PAC, the pairwise function has a fixed parametric form, such as a Gaussian, which limits this techniques generalisability, as a certain choice of function may not suit all problems.  A spatially-varying optical PSF is nearly always smoothly changing, and can be represented as a superposition of a small number of kernels. For this reason, there is no need to introduce an expensive pairwise operation, if a more desirable light-weight method is able to simply interpolate between the kernels. The proposed method is based on this simplification, and enables easier training and faster inference.
Furthermore, being a modification of the convolution itself, PAC cannot take advantage of hardware acceleration for standard convolutions. 

For the sake of completeness one should also mention the arguably most general case of an adaptive convolution, the locally-connected network (LCN) \cite{lcn}. Here, each position uses its own kernel to connect to a region in the feature map. Unfortunately, this flexibility comes at a price and LCNs require a much larger number of parameters than CNNs, which requires more memory, makes them prone to overfitting and harder to train. Furthermore, similarly to pixel-adaptive convolution, these networks cannot use the GPU-level optimization of pure convolutions and thus, are significantly slower to execute than deep CNNs.\newline

\noindent\textbf{Attention.} Finally, a brief description is provided for the attention mechanism, which bares some similarity to the proposed method and has received significant recent interest in computer vision \cite{attention_in_imaging,attention_imaging1,att_unet}. Attention acts on an input vector $\mathbf{v}$ to give an output vector $\mathbf{z}$, and its general form is given by,
\begin{equation}\label{attention}
    \mathbf{z} = f(g(\mathbf{v}),\mathbf{v}),
\end{equation}
where $g$ is a function to generate the attention, and $f$ applies the attention to $\mathbf{v}$. Common forms of $f$ include element-wise multiplication, weighted sum, or concatenation, while $g$ can be a simple linear transformation, a neural network, or even a more complex function. A popular evolution of the method, self-attention, refers to the case where the generator function depends on pairwise function of elements of $\mathbf{v}$. When used as spatial attention, this addresses a limitation of the receptive fields of convolutions, and has thus received use in computer vision in order to capture non-local features \cite{nonlocal_nn}. One method utilised self attention for image recognition \cite{imageselfattention}, and also applied a CoordConv inspired technique of concatenating the coordinates to the feature map.

\section{Proposed Method}\label{sec:proposed}

As has been discussed above, current methods either rely on (indirect) modification of the convolution kernel, on concatenating a coordinate map that acts as additional weight, or on the use of an attention mechanism. The proposed method takes inspiration from these approaches, but condenses them into a lightweight module, which is ideally suited for applications such as static spatially-varying convolution in optical imaging.

\begin{figure}[b]
    \centering
    \includegraphics[width=8cm]{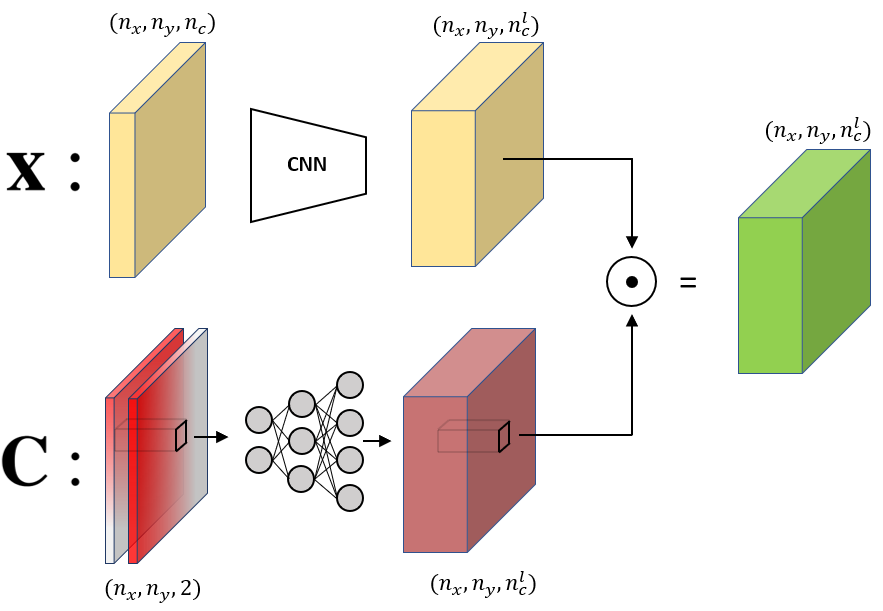}
    \caption{CoordGate. The data, $\mathbf{X}$, and coordinates, $\mathbf{C}$, are fed through a CNN and a MLP respectively, before the Hadamard product is used between the resultant tensors.}
    \label{coordgate}
\end{figure}
In the convolutional \textbf{CoordGate} module (see \cref{coordgate}), the input, $\mathbf{x}\in\mathcal{R} ^{n_x\times n_y\times n_c}$, is first fed through a standard convolutional block, $h$, with the final layer containing $n_c^{l}$ output channels. As discussed previously, these channels correspond to globally applied convolutions. To synthesize a locally-varying convolution from these, the output channels are then multiplied with a gating mask of the same size, somewhat similar to an attention map. However, in contrast to attention that is based on the input signal, here we create the gating mask by taking a static coordinate map, $\mathbf{C}\in\mathcal{R} ^{n_x\times n_y\times 2}$, as in CoordConv, and feeding it through a pixel-wise fully-connected encoding network, $g$, with the last layer containing $n_c^{l}$ neurons. If one is using residual learning, a final $1\times1$ convolutional layer can be used to yield an output with the original number of channels. Denoting an index of the two-dimensional arrays with the vector $\mathbf{i}$, and the channel slice of the resultant vector, $\mathbf{y}$, with $a$, CoordGate is described by, 
\begin{equation}
    \mathbf{y}_{\mathbf{i},a} = \left(h(\mathbf{x})_{\mathbf{i}} \cdot g\left( \mathbf{C}_{\mathbf{i}} \right)\right)_{a}.
\end{equation}
The intuition behind the method is as follows. The convolutional network $h(\mathbf{x})$ can learn a wide range of resultant kernels, and stores $n_c^{l}$ different ones in the channels of its feature map. If the network includes downsampling and upsampling, such as the U-Net, these kernels can be very large and encorperate non-local features. In order to selectively attenuate filters, adopting a different resultant convolution for each pixel, the feature map is multiplied by the channel-wise attention generated from the coordinates. In other words, the feature channels form a basis whose amplitudes at each position are encoded in the gating map. Importantly, the encoding network $g(\mathbf{C}_\mathbf{i})$ is only dependent on the coordinates and once trained, one directly saves the parameters of the gating map. As a result, during inference the only computational overhead compared to a standard convolution is an element-wise multiplication. 

The previously discussed zero-padding boundary effects can also be seen as a position-encoded multiplication operation. A $3\times3$ kernel being passed over an image will include 3 pixels of 0 when at the boundary (assuming it is not in a corner); accordingly, the corresponding pixel in the feature map will have an average relative magnitude of $\frac{2}{3}$ compared to a pixel in the center. This effect propagates inwards with successive convolutions and the proposed method involves a more deliberate and efficient use of this desirable property of multiplication.

Also note that one could, in principle, remove the position decoder network and make the gating map directly trainable. However, this would not only result in a much larger amount of parameters for the network that will complicate training, but more importantly it would ignore the fact that many systems exhibiting a spatially-varying convolution vary smoothly over the input space and hence, can be parameterized. One might further note that, interestingly, if one were to train the map directly, and the kernels within the convolutional layer form a complete base, this method becomes equally expressive as a locally-connected neural network with shared bias term.

\section{Experiments}\label{sec:experiments}

Here the proposed method is implemented to find solutions to a number of problems. Firstly the trivial example of performing 1D spatially-varying convolution is considered, in order to clearly demonstrate the technique's effectiveness compared to common alternatives. Secondly, the method is applied to the practical problem of image deblurring (2D spatially-varying deconvolution).

\subsection{Learning 1D spatially-varying convolutions}\label{1dexperiment}
\cref{SVconvolution} can be rewritten as the following matrix multiplication,
\begin{equation}
    \vec{n} = \mathbf{H} \vec{m},
\end{equation}
where for the 1D case, $\vec{m}$ and $\vec{n}$  are the original and convolved data, and $\mathbf{H}$ is the convolution matrix. Note that in the case of a spatially invariant convolution, $\mathbf{H}$ has the form of a Toeplitz matrix, and can be approximated perfectly by a single-channel convolutional layer with a suitable kernel size.

Here, 10000 normalized uniform random samples of size (30) are generated for $\vec{m}$, before being multiplied with a custom convolution matrix $\mathbf{H}$ to give $\vec{n}$. The task of the network is to predict $\vec{n}$ given $\vec{m}$, thus approximating $\mathbf{H}$. The proposed method is benchmarked against a number of convolutional architectures, demonstrating their limitations. Each network is trained to convergence using the Adam optimizer \cite{adam} (with default parameters) to minimize the mean squared error (MSE). 

\begin{figure*}[htb]
    \centering
    \includegraphics[width=14.5cm]{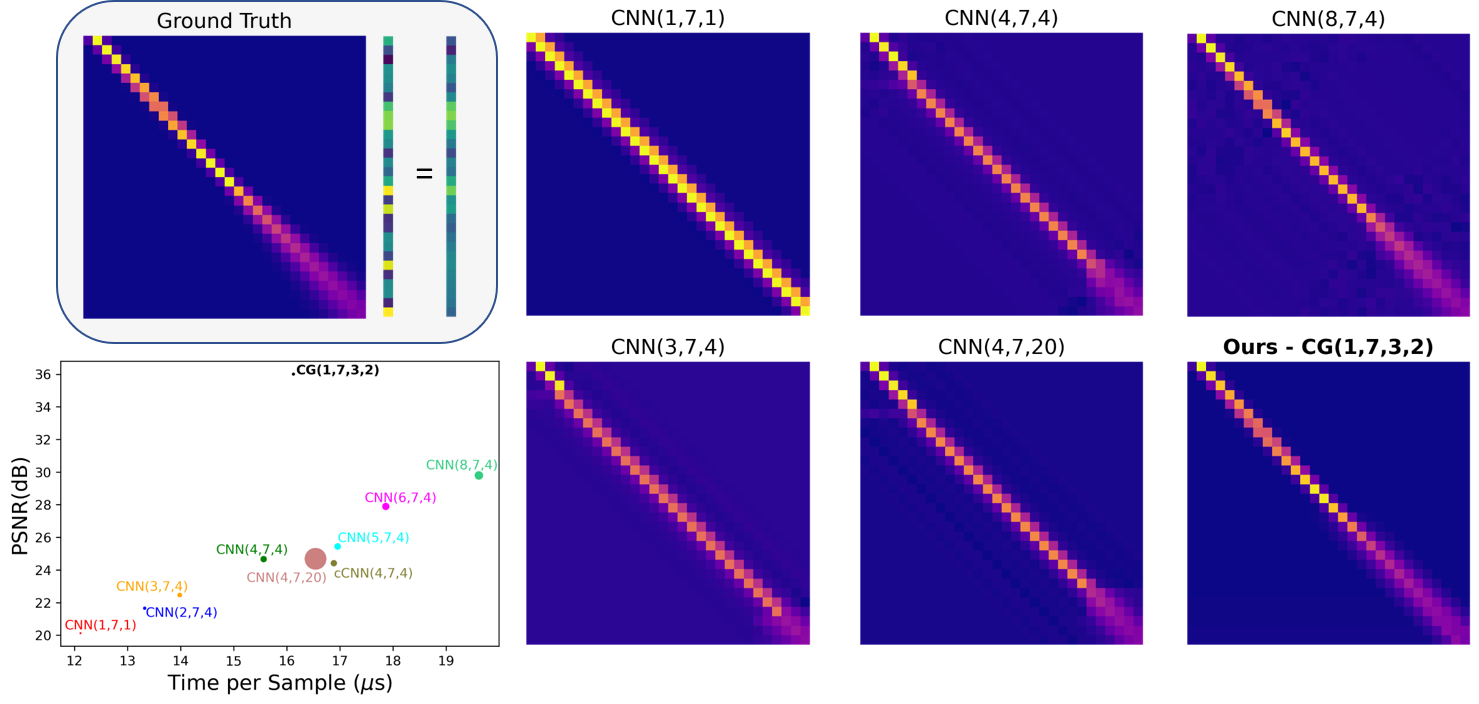}
    \caption{Showing the approximations of the convolution matrix by different models. Also shown is a plot of PSNR against inference time for each model, with the spot size being proportional to the number of parameters in the model.}
    \label{simple_conv_results}
\end{figure*}

A convolutional neural network is denoted by $\textrm{CNN}(L,k,c)$, containing  $L$ convolutional layers with a kernel size of $k$ with $c$ channels, except from the last layer which has 1 channel. A convolutional model utilising CoordConv layers is denoted by $\textrm{cCNN}(L,k,c)$ and the CoordGate architecture is denoted by $\textrm{CG}(L,k,c,p)$, where $L,k,c$ describe the parameters of the convolutional network, and $p$ describes the number of fully connected layers to encode the coordinates, each with $c$ nodes. The custom convolution matrix consists of a Gaussian for each pixel, with varying offset, $\delta(x')$, and standard deviation, $\sigma(x')$, as described by the following equation and shown in shown in \cref{simple_conv_results},

\begin{multline}
    \mathbf{H}(x,x') = e^{\frac{-(x-\delta (x'))^2}{2\sigma (x')^2}}, \:\:\:\: \\(\delta(x'),\sigma(x'))=
    \begin{cases}
      (1-\frac{x}{15},0.5), & \text{if}\ x<15 \\
      (0,0.5(2-\frac{x}{15}) + 2(1-\frac{x}{15})), & \text{otherwise}
    \end{cases}.
\end{multline}

\noindent\textbf{Discussion.}
The graph in \cref{simple_conv_results} displays the peak signal to noise ratio (PSNR), defined as $10 \times\log_{10}(\frac{1}{MSE})$, against the inference time for each model. Beginning with the CNNs, the simple case of a single convolutional layer is intuitive, with the network learning the mean kernel. One then sees an increase in performance with successive layers added. This is explained by the padding effects described earlier, and is seen between the CNN(3,7,4) and CNN(4,7,4) plots, where the boundary is moving down the image, allowing the network to learn spatially variant information in that region, whilst it learns the mean for the rest. This propagation of spatial information is also seen for CNN(8,7,4). The comparison of CNN(4,7,4) and CNN(4,7,20) demonstrates that adding more channels does not help the model to learn spatially varying features. The CoordConv model possesses no benefit over a standard convolutional model in this task - in fact, studying the convolutional kernel, the model tries to discard the effect of coordinate layer. On the contrary, the model utilising CoordGate learns the convolution matrix using one single convolutional layer, with signicantly less inference time and a fraction of the number of parameters required by a convolutional model. The convolution matrix can be thought of as a superposition between 3 gaussian kernels: ($\delta = 1$, $\sigma = 0.5$),($\delta = 0$, $\sigma = 0.5$) and ($\delta = 0$, $\sigma = 2$), so it is intuitive that the CoordGate model required 3 channels in order to interpolate between them. Also tested was the case when instead of coordinates, the static variable was initialised as random numbers ($ \mathcal{U}(-1,1)$), which was found to not converge, highlighting the use of the coordinates. 

Futhermore, the CoordGate method scales much better with increasing sample size, or with reduced kernel size, as it does not require subsequent convolutions to encode the position.

\subsection{Image deblurring}
Image deblurring is an example of a spatially-variant deconvolution. Of the different possible types of blur, many are dynamic - that is, they may vary from image to image - including things like motion blur. Here we consider a static blur, solely originating from imperfections in the imaging system. This is a problem that has received interest in fields such as microscopy  \cite{multiweinernet,microscopydeconv2} and astronomy \cite{astrodeconv,astrodeconv2}, and an effective image deblurrer could also be used to sharpen images from an imperfect relay imaging system \cite{HPLSE}. 

The U-Net is well suited to spatially-varying deconvolution due to its ability to synthesise resultant kernels with wide receptive fields, which helps with two problems; firstly, deconvolution typically requires a much wider receptive field than convolution, due to a more non-local resultant kernel (which theoretically spans the whole image due to mixing, but can be approximated with a smaller kernel). Secondly, as has already been discussed, in order to capture spatially varying features throughout the image, the zero-padding defect must be propagated to the center. This second condition requires the U-Net to be very deep, as is seen in \cref{boundary}, which adds a large number of parameters (as typically the number of channels increases with depth). Here, we hypothesise that adding CoordGate to the model will allow for a more efficient extraction of spatially-varying behaviour, and thus allow a much shallower U-Net to be able to achieve equal performance, with a fraction of the parameters.  

\begin{figure*}[!htb]
    \centering
    \includegraphics[width=14.5cm]{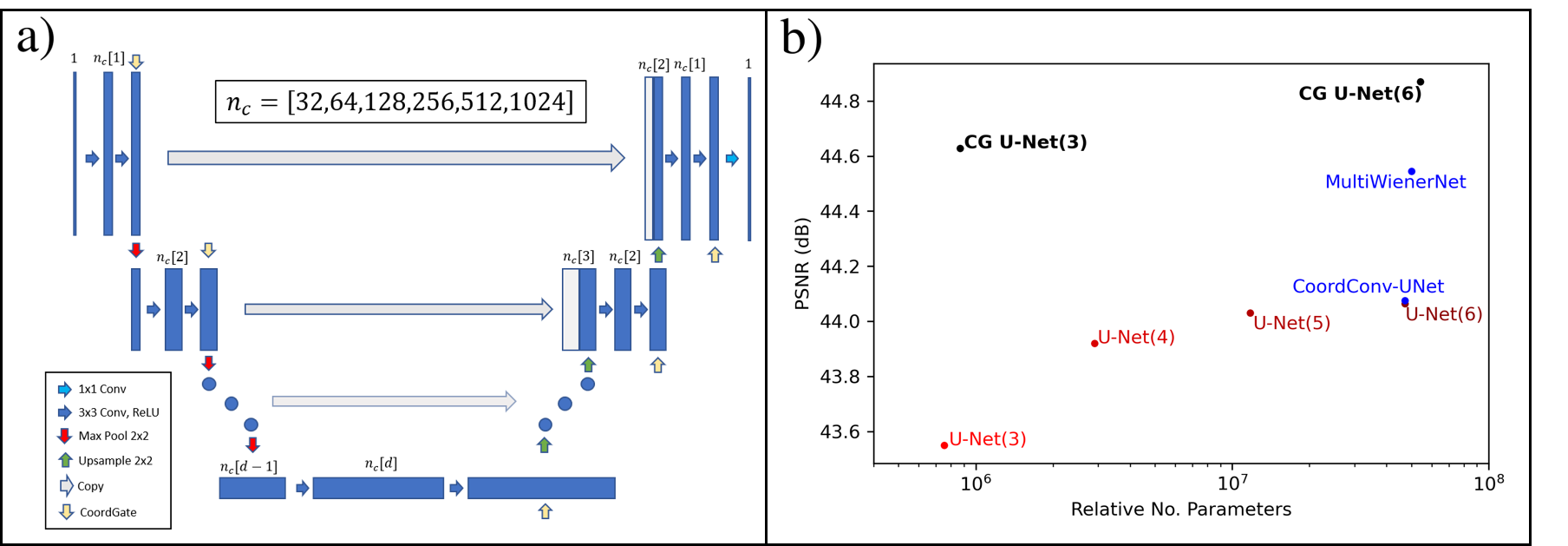}
    \caption{(a): The backbone U-Net architecture. A model with depth, $d$, has $n_c[d]$ channels in its deepest layer. The yellow CoordGate arrows are added for the CG U-Net($d$) models. (b): A plot of PSNR against the logarithm of the number of parameters, demonstrating the advantage of adding CoordGate to the U-Net architecture. Also included are the CoordConv-UNet and MultiWienerNet models.}
    \label{fig:deblur_result}
\end{figure*}

Multiple U-Nets are used, and an architecture with depth $d$ is denoted as U-Net$(d)$. The form of these models is seen  in \cref{fig:deblur_result}(a). In every step there are two $3\times 3$ convolutions with the rectified linear unit activation function (ReLU) applied \cite{relu}. To test CoordGate's effectiveness in this scenario, it is added to both the most shallow and deep U-Nets at each down-or-up sampling point.  Furthermore, we also compare CoordGate to CoordConv-UNet\cite{el2021coordconv}, and a state-of-the-art method in this setting of image deblurring from a static point spread function, named MultiWienerNet  \cite{multiweinernet}. This technique first requires the measurement of the PSF at a number of locations across the sensor (which represents an inconvenience not required for CoordGate). Then Wiener deconvolution is performed each of these PSFs, resulting in a number of feature maps that are relatively unblurred in the positions of where each PSF was measured. Finally, a U-Net is used to combine and refine these maps to give the prediction. One notes that the MultiWienerNet was developed in the setting where the PSF varies very quickly (to be used with compressed sensing \cite{miniscope}), whereas here we consider a PSF of a typical lens. Both CoordConv-UNet and MultiWienerNet were implemented with the deepest U-Net architecture, U-Net(6).

\begin{figure*}[!htb]
    \centering
    \includegraphics[width=14.5cm]{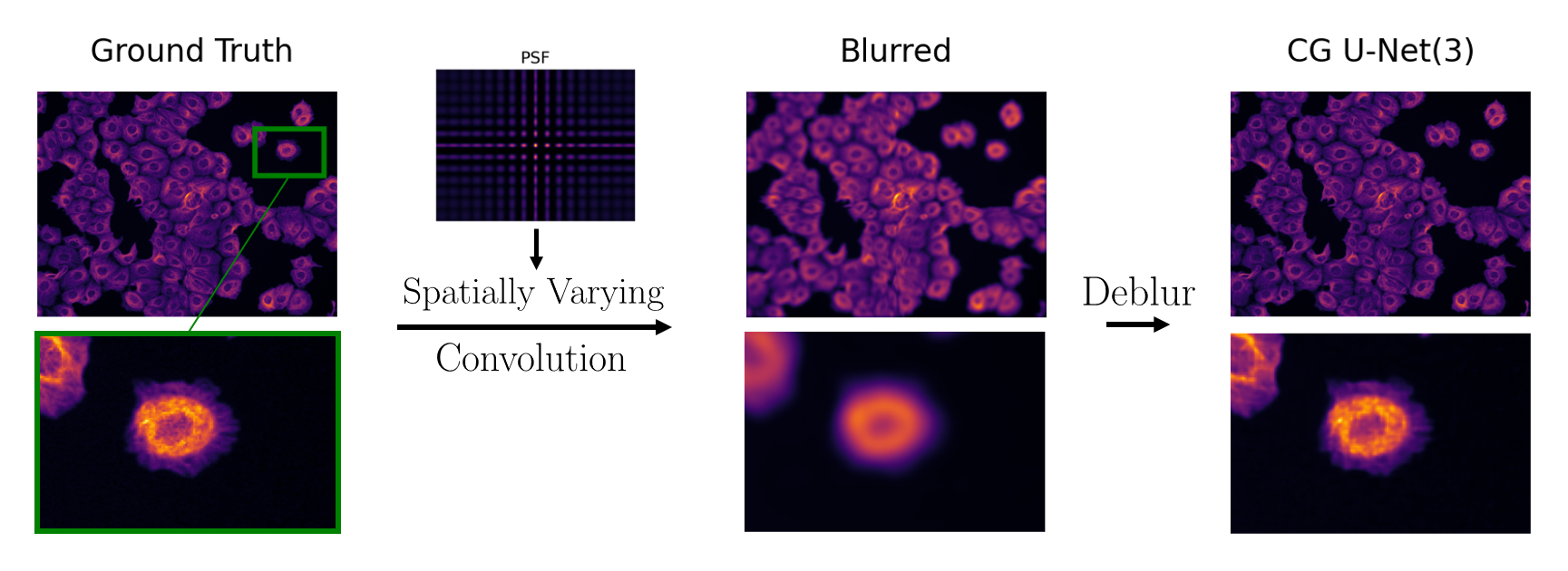}
    \caption{An example showing the blurring and subsequent de-blurring process using our CoordGate technique. }
    \label{fig:deblur_example}
\end{figure*}

Here we utilise the database of microscopy images of live cells collected from multiple sources \cite{microscopydata1,microscopydata2}. A synthetic PSF was applied to each of the 20000 samples, with defocus increasing towards the edge of the image to try and simulate a realistic lens, which is seen in \cref{fig:deblur_example}. The job of a model, is to recover the original unblurred image from its blurred counterpart. Each model was trained with the Adam optimizer for 600 epochs. The initial learning rate was 0.001, and was halved if the validation loss didn't decrease over 20 epochs. An example of a reconstruction is shown in \cref{fig:deblur_example}.\\

\noindent\textbf{Discussion.} \cref{fig:deblur_result}((b) shows the validation PSNR of each trained model against the number of parameters. Comparing the different U-Net architectures, one sees that the PSNR increases with depth, as expected due to the increased capacity to represent spatially-varying features that has been described previously. The form of the PSF includes a region of focus in the center, which explains why $\textrm{U-Net}(6)$ performs only marginally better than $\textrm{U-Net}(5)$, as the latter model can still represent this section well with the average kernel. 

The addition of CoordGate causes a large increase in performance and efficiency. In fact, upon the implementation of CoordGate, the most shallow model ($\textrm{CG U-Net}(3)$) is able to outperform the deepest normal model ($\textrm{U-Net}(6)$), despite the fact that it contains 60$\times$ less parameters. The deeper model has much more capacity to synthesise bigger kernels, which gives it a natural advantage, and thus the fact that it is outperformed, shows that CoordGate is superior in learning the spatially-varying features. This claim is supported by studying the average loss over all examples, which shows a significantly lower loss close to the center of the image for CoordGate. Finally, one notes that $\textrm{CG U-Net}(6)$ performs better than $\textrm{CG U-Net}(3)$, which is explainable by the fact that it can synthesise wider kernels, therefore better approximating the true global deconvolution kernels. 

It was found that the CoordConv-UNet model performed nearly identically to the base U-Net(6) model, suggesting that the concatenation operation doesn't provide the coordinate information to the network in an optimum way. The incorporation of Wiener deconvolution in MultiWienerNet resulted in a significant performance increase over the base U-Net. However, it must be remembered that this model is given more information, in the form of some prior measurements of the PSF. Even so, it is outperformed by the shallower CoordGate model, despite containing significantly more parameters. Experiments with varying the size of the fully connected layer are included in the supplemental material, along with additional metrics.

\section{Conclusion}\label{sec:conclusions}

This paper has introduced a new method, CoordGate, to allow for more efficient and accurate spatially-varying convolution and deconvolution. The technique works by multiplying the output of a CNN by a gating map, that is generated by an pixel-wise coordinate-encoding network, thereby selectively attenuating the resultant convolutional kernels for each pixel. CoordGate can be seen as lightweight in two ways. Firstly, the implementation itself adds minimal parameters to the existing model. Secondly, the addition of CoordGate allows a much simpler backbone network to achieve a superior performance than that of a more relatively complex model, as has been proven in the experiments.

CoordGate's utility was first verified on the simple case of 1D convolution, before being applied to a more challenging problem of removing a spatially-varying blur caused by a lens. In the latter, adding CoordGate modules to a shallow U-Net architecture enabled it to achieve a higher accuracy than a much deeper U-Net architecture, despite having almost two order of magnitude less parameters. For this problem, CoordGate also outperformed two recent methods MultiWienerNet and CoordConv-UNet in terms of outright accuracy and efficiency. Further work should involve the implementation of the CoordGate module to different models and problems; in particular the authors plan to use CoordGate to mitigate imperfections in the relay system of a snapshot compressive imaging apparatus \cite{HPLSE}.

\section*{Acknowledgements}
\textit{We would like to acknowledge the useful discussions with Dr. Ramy Aboushelbaya and the rest of Professor Peter Norreys' group, as well as the members of the DOLPHIN group at LMU's Centre for Advanced Laser Applications. This work was supported by the Independent Junior Research Group ``Characterization and control of high-intensity laser pulses for particle acceleration", DFG Project No.~453619281. We would also like to acknowledge UKRI-STFC grant ST/V001655/1.}

\onecolumngrid
\newpage

\section{Appendix}
\twocolumngrid

This section provides some additional results and metrics.

\subsection*{Training Curves}
Training curves for the results displayed in Fig. 4b. Each model was trained with the Adam optimizer with default initial parameters, and the learning rate was set to half if the validation loss did not decrease for 10 epochs. 

\begin{figure}[h]
    \centering
\includegraphics[width=\linewidth]{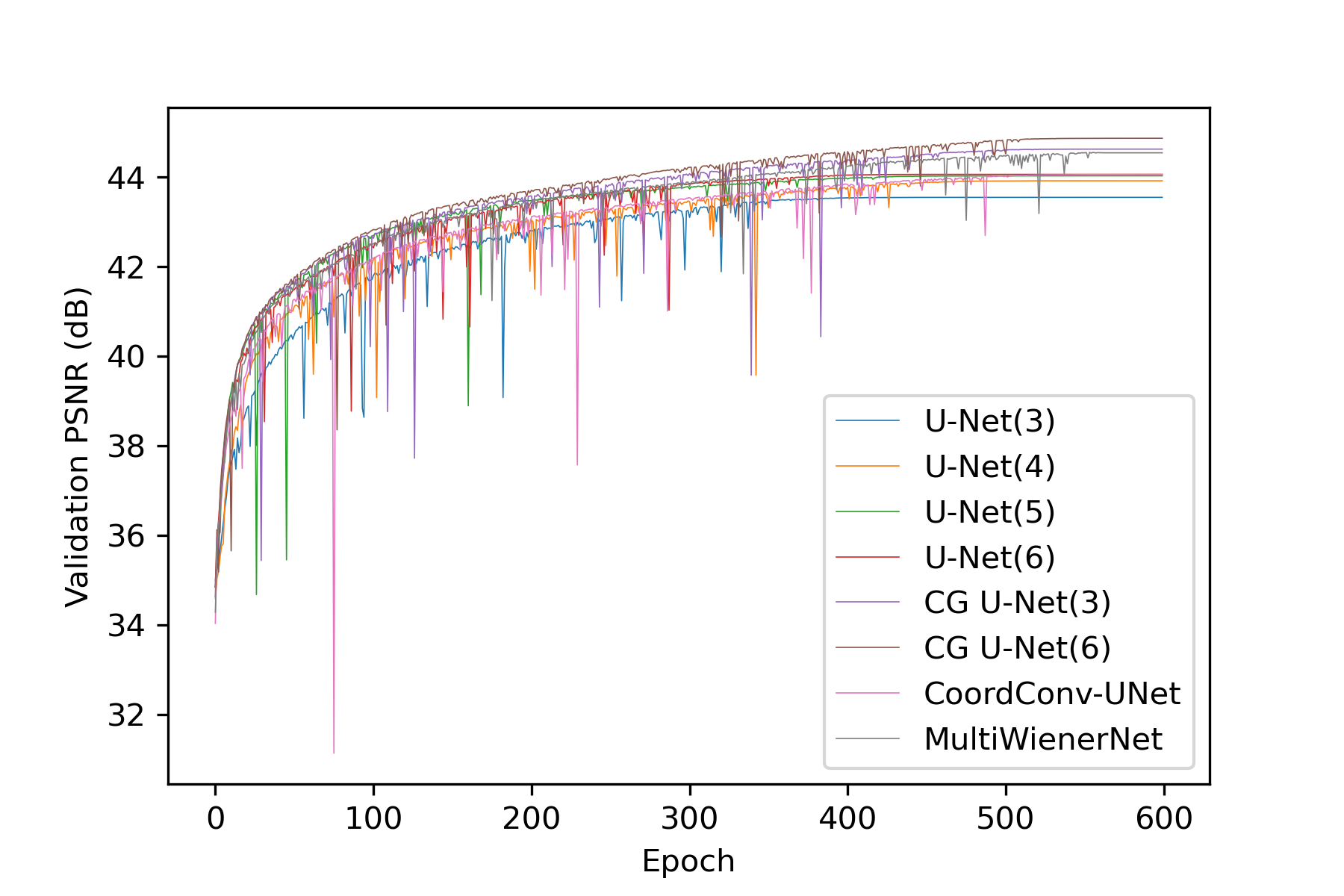}
    \caption{Curves showing the PSNR evaluated on the validation dataset during training, for each model.}
    \label{fig:traincurve}
\end{figure}

\subsection*{SSIM Results}
The structural similarity index measure (SSIM) for each trained model, evaluated on the test set. We see that these results follow the trend of the PSNR.

\begin{figure}[h]
    \centering
\includegraphics[width=\linewidth]{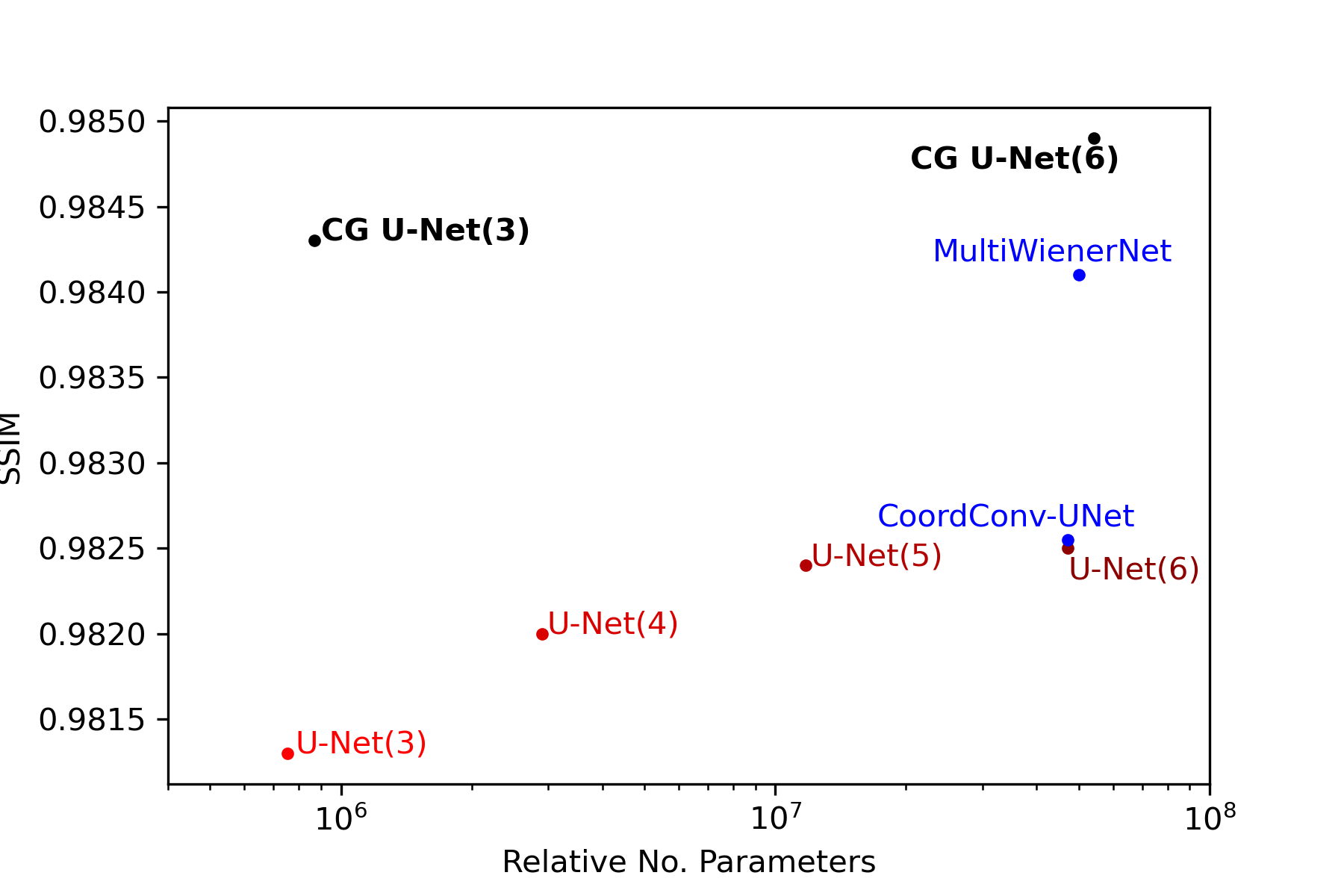}
    \caption{The SSIM of each trained models predictions, evaluated on the test dataset.}
    \label{fig:SSIM}
\end{figure}

\section*{Equivalency to Locally Connected Layer}

In the situation that the kernels within the convolutional layer form a complete basis, the CoordGate module is equally expressive as a locally-connected layer. To see this, let's take a simple example case of a $3\times3$ convolution, where a complete basis is for instance formed by 9 ``pixel-basis" filters of the form:
\begin{equation}
    \begin{bmatrix} 1&0&0\\0&0&0\\0&0&0 \end{bmatrix}, 
    \begin{bmatrix} 0&1&0\\0&0&0\\0&0&0 \end{bmatrix},
     \cdots, \begin{bmatrix} 0&0&0\\0&0&0\\0&0&1 \end{bmatrix}
\end{equation}

CoordGate encodes position into a gating map, to amplify or dampen the individual filters at each position of the convolutional feature map. For instance, at two adjacent spatial positions, the gating map tensor could have the values [1,2,1,1,1,1,1,1,1] and [0,2,1,1,1,1,1,1,1]. After the Hadamard product is taken between the gating map and the feature map, summing over the channel dimension will result in combined filters: 

\begin{equation}
    \begin{bmatrix} 1&2&1\\1&1&1\\1&1&1 \end{bmatrix}, 
    \begin{bmatrix} 0&2&1\\1&1&1\\1&1&1 \end{bmatrix}.
\end{equation}

So we have broken the weight sharing inherent to convolutional layers and instead have individual filters for the receptive field of each neuron. This is exactly what a locally-connected neural network does. If we were to add an additional gating map this operation would be identical to a locally-connected layer; without it all neurons have a shared bias term.


\begin{thebibliography}{0}%
\makeatletter
\providecommand \@ifxundefined [1]{%
 \@ifx{#1\undefined}
}%
\providecommand \@ifnum [1]{%
 \ifnum #1\expandafter \@firstoftwo
 \else \expandafter \@secondoftwo
 \fi
}%
\providecommand \@ifx [1]{%
 \ifx #1\expandafter \@firstoftwo
 \else \expandafter \@secondoftwo
 \fi
}%
\providecommand \natexlab [1]{#1}%
\providecommand \enquote  [1]{``#1''}%
\providecommand \bibnamefont  [1]{#1}%
\providecommand \bibfnamefont [1]{#1}%
\providecommand \citenamefont [1]{#1}%
\providecommand \href@noop [0]{\@secondoftwo}%
\providecommand \href [0]{\begingroup \@sanitize@url \@href}%
\providecommand \@href[1]{\@@startlink{#1}\@@href}%
\providecommand \@@href[1]{\endgroup#1\@@endlink}%
\providecommand \@sanitize@url [0]{\catcode `\\12\catcode `\$12\catcode
  `\&12\catcode `\#12\catcode `\^12\catcode `\_12\catcode `\%12\relax}%
\providecommand \@@startlink[1]{}%
\providecommand \@@endlink[0]{}%
\providecommand \url  [0]{\begingroup\@sanitize@url \@url }%
\providecommand \@url [1]{\endgroup\@href {#1}{\urlprefix }}%
\providecommand \urlprefix  [0]{URL }%
\providecommand \Eprint [0]{\href }%
\providecommand \doibase [0]{https://doi.org/}%
\providecommand \selectlanguage [0]{\@gobble}%
\providecommand \bibinfo  [0]{\@secondoftwo}%
\providecommand \bibfield  [0]{\@secondoftwo}%
\providecommand \translation [1]{[#1]}%
\providecommand \BibitemOpen [0]{}%
\providecommand \bibitemStop [0]{}%
\providecommand \bibitemNoStop [0]{.\EOS\space}%
\providecommand \EOS [0]{\spacefactor3000\relax}%
\providecommand \BibitemShut  [1]{\csname bibitem#1\endcsname}%
\let\auto@bib@innerbib\@empty
\end{thebibliography}%


\begin{thebibliography}{10}

\bibitem{boundary_effects}
O.~S. Kayhan and J.~C. van Gemert, ``On translation invariance in cnns:
  Convolutional layers can exploit absolute spatial location,'' in {\em 2020
  IEEE/CVF Conference on Computer Vision and Pattern Recognition (CVPR)}, (Los
  Alamitos, CA, USA), pp.~14262--14273, IEEE Computer Society, jun 2020.

\bibitem{coordconv}
R.~Liu, J.~Lehman, P.~Molino, F.~Petroski~Such, E.~Frank, A.~Sergeev, and
  J.~Yosinski, ``An intriguing failing of convolutional neural networks and the
  coordconv solution,'' in {\em Advances in Neural Information Processing
  Systems} (S.~Bengio, H.~Wallach, H.~Larochelle, K.~Grauman, N.~Cesa-Bianchi,
  and R.~Garnett, eds.), vol.~31, Curran Associates, Inc., 2018.

\bibitem{boundary_effects2}
M.~A. Islam, S.~Jia, and N.~D.~B. Bruce, ``How much position information do
  convolutional neural networks encode?,'' 2020.

\bibitem{unet}
O.~Ronneberger, P.~Fischer, and T.~Brox, ``{U-Net: Convolutional Networks for
  Biomedical Image Segmentation},'' {\em arXiv}, 2015.

\bibitem{wang2020solo}
X.~Wang, T.~Kong, C.~Shen, Y.~Jiang, and L.~Li, ``Solo: Segmenting objects by
  locations,'' in {\em Computer Vision--ECCV 2020: 16th European Conference,
  Glasgow, UK, August 23--28, 2020, Proceedings, Part XVIII 16}, pp.~649--665,
  Springer, 2020.

\bibitem{imageselfattention}
H.~Zhao, J.~Jia, and V.~Koltun, ``Exploring self-attention for image
  recognition,'' in {\em 2020 IEEE/CVF Conference on Computer Vision and
  Pattern Recognition (CVPR)}, pp.~10073--10082, 2020.

\bibitem{sitzmann2019deepvoxels}
V.~Sitzmann, J.~Thies, F.~Heide, M.~Nie{\ss}ner, G.~Wetzstein, and
  M.~Zollhofer, ``Deepvoxels: Learning persistent 3d feature embeddings,'' in
  {\em Proceedings of the IEEE/CVF Conference on Computer Vision and Pattern
  Recognition}, pp.~2437--2446, 2019.

\bibitem{el2021coordconv}
R.~{El Jurdi}, C.~Petitjean, P.~Honeine, and F.~Abdallah, ``Coordconv-unet:
  Investigating coordconv for organ segmentation,'' {\em IRBM}, vol.~42, no.~6,
  pp.~415--423, 2021.

\bibitem{svcnn}
Y.~Dai, T.~Jin, Y.~Song, S.~Sun, and C.~Wu, ``Convolutional neural network with
  spatial-variant convolution kernel,'' {\em Remote Sensing}, vol.~12, no.~17,
  2020.

\bibitem{uselis2020localized}
A.~Uselis, M.~Luko{\v{s}}evi{\v{c}}ius, and L.~Stasytis, ``Localized
  convolutional neural networks for geospatial wind forecasting,'' {\em
  Energies}, vol.~13, no.~13, p.~3440, 2020.

\bibitem{pixeladaptiveconvolution}
H.~Su, V.~Jampani, D.~Sun, O.~Gallo, E.~Learned-Miller, and J.~Kautz,
  ``Pixel-adaptive convolutional neural networks,'' in {\em Proceedings of the
  IEEE/CVF Conference on Computer Vision and Pattern Recognition},
  pp.~11166--11175, 2019.

\bibitem{lcn}
J.~Bruna, W.~Zaremba, A.~Szlam, and Y.~Lecun, ``Spectral networks and locally
  connected networks on graphs,'' in {\em International Conference on Learning
  Representations (ICLR2014), CBLS, April 2014}, 2014.

\bibitem{attention_in_imaging}
M.-H. Guo, T.-X. Xu, J.-J. Liu, Z.-N. Liu, P.-T. Jiang, T.-J. Mu, S.-H. Zhang,
  R.~R. Martin, M.-M. Cheng, and S.-M. Hu, ``Attention mechanisms in computer
  vision: A survey,'' {\em Computational Visual Media}, vol.~8, no.~3,
  pp.~331--368, 2022.

\bibitem{attention_imaging1}
Y.~Zhang, K.~Li, K.~Li, L.~Wang, B.~Zhong, and Y.~Fu, ``Image super-resolution
  using very deep residual channel attention networks,'' in {\em Computer
  Vision -- ECCV 2018} (V.~Ferrari, M.~Hebert, C.~Sminchisescu, and Y.~Weiss,
  eds.), (Cham), pp.~294--310, Springer International Publishing, 2018.

\bibitem{att_unet}
O.~Oktay, J.~Schlemper, L.~L. Folgoc, M.~Lee, M.~Heinrich, K.~Misawa, K.~Mori,
  S.~McDonagh, N.~Y. Hammerla, B.~Kainz, B.~Glocker, and D.~Rueckert,
  ``Attention u-net: Learning where to look for the pancreas,'' in {\em Medical
  Imaging with Deep Learning}, 2018.

\bibitem{nonlocal_nn}
X.~Wang, R.~Girshick, A.~Gupta, and K.~He, ``Non-local neural networks,'' in
  {\em Proceedings of the IEEE conference on computer vision and pattern
  recognition}, pp.~7794--7803, 2018.

\bibitem{adam}
D.~P. Kingma and J.~Ba, ``{Adam: A Method for Stochastic Optimization},'' {\em
  3rd International Conference on Learning Representations, ICLR 2015 -
  Conference Track Proceedings}, 12 2014.

\bibitem{multiweinernet}
K.~Yanny, K.~Monakhova, R.~W. Shuai, and L.~Waller, ``Deep learning for fast
  spatially varying deconvolution,'' {\em Optica}, vol.~9, pp.~96--99, Jan
  2022.

\bibitem{microscopydeconv2}
B.~Toader, J.~Boulanger, Y.~Korolev, M.~O. Lenz, J.~Manton, C.-B.
  Sch{\"o}nlieb, and L.~Mure{\c s}an, ``Image reconstruction in {Light-Sheet}
  microscopy: Spatially varying deconvolution and mixed noise,'' {\em J Math
  Imaging Vis}, vol.~64, pp.~968--992, June 2022.

\bibitem{astrodeconv}
T.~Lauer, ``Deconvolution with a spatially-variant psf,'' {\em Proc SPIE},
  vol.~4847, 08 2002.

\bibitem{astrodeconv2}
{Farrens, S.}, {Ngol\`e Mboula, F. M.}, and {Starck, J.-L.}, ``Space variant
  deconvolution of galaxy survey images,'' {\em A\&A}, vol.~601, p.~A66, 2017.

\bibitem{HPLSE}
S.~Howard, J.~Esslinger, R.~H.~W. Wang, P.~Norreys, and A.~Döpp,
  ``Hyperspectral compressive wavefront sensing,'' {\em High Power Laser
  Science and Engineering}, vol.~11, p.~e32, 2023.

\bibitem{relu}
A.~F. Agarap, ``Deep learning using rectified linear units (relu),'' {\em arXiv
  preprint arXiv:1803.08375}, 2018.

\bibitem{miniscope}
K.~Yanny, N.~Antipa, W.~Liberti, S.~Dehaeck, K.~Monakhova, F.~L. Liu, K.~Shen,
  R.~Ng, and L.~Waller, ``Miniscope3d: optimized single-shot miniature 3d
  fluorescence microscopy,'' {\em Light, Science \& Applications}, vol.~9,
  2020.

\bibitem{microscopydata1}
V.~Ljosa, K.~Sokolnicki, and A.~Carpenter, ``Annotated high-throughput
  microscopy image sets for validation,'' {\em Nature methods}, vol.~9, p.~637,
  06 2012.

\bibitem{microscopydata2}
Y.~Zhang, Y.~Zhu, E.~Nichols, Q.~Wang, S.~Zhang, C.~Smith, and S.~Howard, ``A
  poisson-gaussian denoising dataset with real fluorescence microscopy
  images,'' in {\em 2019 IEEE/CVF Conference on Computer Vision and Pattern
  Recognition (CVPR)}, (Los Alamitos, CA, USA), pp.~11702--11710, IEEE Computer
  Society, jun 2019.

\end{thebibliography}
\end{document}